\documentclass[sigconf]{acmart}

\usepackage{amsmath}
\usepackage{amsthm}
\usepackage{enumitem}
\usepackage[most]{tcolorbox}
\usepackage[ruled,linesnumbered]{algorithm2e}
\usepackage{hhline}
\usepackage{arydshln}
\usepackage{color, colortbl}
\usepackage{bm}
\usepackage{hhline}
\usepackage{dblfloatfix}

\def\Plus{\texttt{+}}
\def\Minus{\texttt{-}}

\usepackage{tikz}
\usetikzlibrary{arrows, automata}
\usetikzlibrary{decorations.pathreplacing,angles,quotes,arrows.meta,%
                backgrounds,
                calligraphy,
                positioning,calc}
\newcommand{\tikzmark}[1]{\tikz[overlay,remember picture] \node (#1) {};}

\newcommand*{\AddNote}[4]{%
    \begin{tikzpicture}[overlay, remember picture]
        \draw [decoration={brace,amplitude=0.5em},decorate, thick,blue]
            ($(#3)!(#1.north)!($(#3)-(0,1)$)$) --  
            ($(#3)!(0,2.6)!($(#3)-(0,1)$)$)
                node [align=center, text width=2.5cm, pos=0.5, anchor=west] {#4};
    \end{tikzpicture}
}%

\pgfdeclaredecoration{lightning bolt}{draw}{
\state{draw}[width=\pgfdecoratedpathlength]{
  \pgfpathmoveto{\pgfpointorigin}%
  \pgfpathlineto{\pgfpoint{\pgfdecoratedpathlength*0.6}%
    {-\pgfdecoratedpathlength*.1}}%
  \pgfpathlineto{\pgfpoint{\pgfdecoratedpathlength*0.55}{0pt}}%
  \pgfpathlineto{\pgfpoint{\pgfdecoratedpathlength}{0pt}}%
  \pgfpathlineto{\pgfpoint{\pgfdecoratedpathlength*0.4}%
    {\pgfdecoratedpathlength*.1}}%
  \pgfpathlineto{\pgfpoint{\pgfdecoratedpathlength*0.45}{0pt}}%
  \pgfpathclose%
}%
}

\usepackage{subcaption}
\DeclareMathOperator*{\argmin}{arg\,min}

\let\oldnl\nl
\newcommand{\nonl}{\renewcommand{\nl}{\let\nl\oldnl}}
\newcommand\ci{\perp\!\!\!\perp}

\makeatletter
\newtheorem*{rep@theorem}{\rep@title}
\newcommand{\newreptheorem}[2]{%
\newenvironment{rep#1}[1]{%
 \def\rep@title{#2 \ref{##1}}%
 \begin{rep@theorem}}%
 {\end{rep@theorem}}}
\makeatother

\definecolor{Gray}{gray}{0.93}

\newtheorem{assumptions1}{Assumption}
\newtheorem{theorem1}{Theorem}
\newtheorem{lemma1}{Lemma}
\newtheorem{proposition1}{Proposition}
\newtheorem{definition1}{Definition}
\newreptheorem{theorem}{Theorem}
\newreptheorem{lemma}{Lemma}
\newreptheorem{proposition}{Proposition}

\usepackage{bbm}

\AtBeginDocument{%
  \providecommand\BibTeX{{%
    \normalfont B\kern-0.5em{\scshape i\kern-0.25em b}\kern-0.8em\TeX}}}

\acmConference[]{}{}{}

\begin{document}

\title{Identifying Patient-Specific Root Causes of Disease}

\author{Eric V. Strobl \& Thomas A. Lasko}

\renewcommand{\shortauthors}{Eric V. Strobl, Thomas A. Lasko}

\begin{abstract}
Complex diseases are caused by a multitude of factors that may differ between patients. As a result, hypothesis tests comparing all patients to all healthy controls can detect many significant variables with inconsequential effect sizes. A few highly predictive root causes may nevertheless generate disease \textit{within each patient}. In this paper, we define \textit{patient-specific root causes} as variables subject to exogenous ``shocks'' which go on to perturb an otherwise healthy system and induce disease. In other words, the variables are associated with the exogenous errors of a structural equation model (SEM), and these errors predict a downstream diagnostic label. We quantify predictivity using sample-specific Shapley values. This derivation allows us to develop a fast algorithm called Root Causal Inference for identifying patient-specific root causes by extracting the error terms of a linear SEM and then computing the Shapley value associated with each error. Experiments highlight considerable improvements in accuracy because the method uncovers root causes that may have large effect sizes at the individual level but clinically insignificant effect sizes at the group level. An R implementation is available at github.com/ericstrobl/RCI. 
\end{abstract}





\maketitle

\section{Introduction}

\begin{figure}
\centering
\begin{tikzpicture}[scale=1.0, shorten >=1pt,auto,node distance=2.8cm, semithick,
  inj/.pic = {\draw (0,0) -- ++ (0,2mm) 
                node[minimum size=2mm, fill=red!60,above] {}
                node[draw, semithick, minimum width=2mm, minimum height=5mm,above] (aux) {};
              \draw[thick] (aux.west) -- (aux.east); 
              \draw[thick,{Bar[width=2mm]}-{Hooks[width=4mm]}] (aux.center) -- ++ (0,4mm) coordinate (-inj);
              }]
                    
\tikzset{vertex/.style = {inner sep=0.4pt}}
\tikzset{edge/.style = {->,> = latex'}}
 
\node[vertex] (1) at  (0,0) {$X_1$};
\node[vertex] (2) at  (1.5,0) {$X_2$};
\node[vertex] (3) at  (3,0.5) {$X_3$};
\node[vertex] (4) at  (3,-0.5) {$X_4$};
\node[vertex] (5) at  (4.5,0.5) {$X_5$};
\node[vertex] (6) at  (4.5,-0.5) {$X_6$};
\node[vertex] (7) at  (6,0) {$D$};

\fill [orange, decoration=lightning bolt, decorate] (1.5,0.25) -- ++ (0.75,0.75);

\draw[edge] (1) to (2);
\draw[edge] (2) to (3);
\draw[edge] (2) to (4);
\draw[edge] (3) to (5);
\draw[edge] (4) to (6);
\draw[edge] (5) to (7);
\draw[edge] (6) to (7);
\end{tikzpicture}

\caption{The lightning bolt corresponds to a ``shock'' to the causal system at $X_2$. This perturbed version of $X_2$ in turn affects many downstream variables, such as $\{ X_3, X_4\}$, ultimately causing symptoms $\{X_5, X_6\}$ and a diagnosis $D$.} \label{fig_shock}
\end{figure}
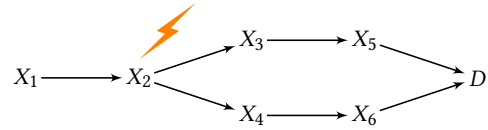

Causal inference refers to the process of statistically inferring causation from data. Scientists usually infer causation from randomized controlled trials (RCTs). RCTs however cannot distinguish between a cause and a \textit{root cause}, or an initial change that induces disease. A root cause intuitively corresponds to an exogenous ``shock'' in an otherwise normal system that leads to multiple downstream effects ultimately causing symptoms and then a diagnostic label (Figure \ref{fig_shock}).

This intuitive explanation of a root cause nevertheless lacks a rigorous definition which in turn hinders the development of algorithms designed to detect root causes. Creating a principled framework is critical, in particular for the discovery of the root causes of \textit{complex disease}, or illness believed to be induced by a multitude of factors that may differ between patients even within the same diagnostic category. Each patient may have a unique combination of multiple root causes, so we must define and then devise a way to more specifically identify sample or \textit{patient-specific root causes} -- with the ultimate goal of course to find treatment targets that reverse the initial shocks to the system. Failing to do so and only treating the downstream effects more proximal to the diagnosis, such as $X_4$ in Figure \ref{fig_shock}, may not fully eliminate the disease.

Existing methods unfortunately cannot recover patient-specific root causes, such as $X_2$. The problem is challenging because the diagnosis may lie far downstream from a root cause. Even well-developed approaches like dependence testing or predictive modeling fail for this reason. For example, dependence testing recovers all of the variables other than the diagnosis $D$ in Figure \ref{fig_shock} (under mild conditions), and even well-specified models predicting the diagnosis only recover $\{X_5,X_6\}$. Most algorithms specialized towards causal discovery also assume that the diagnosis \textit{perfectly} indexes changes in the underlying causal process, even though the diagnosis is actually an \textit{imperfect} index made by clinical judgement based on downstream symptoms \cite{Wang18,Ghoshal19,Huang20,Mooij20,Budhathoki21}. Only one method attempts to define a sample-specific root cause as a conditional outlier, but this definition is more suitable for engineering applications like factory machinery rather than medicine because not all outliers predictably induce disease and not all diseases are induced by outliers \cite{Janzing19}. Existing methods therefore do not actually target patient-specific root causes. See Section \ref{sec_RW} for a more comprehensive review of related work.

\begin{tcolorbox}[breakable,enhanced,frame hidden]
The above issues motivate the following contributions in this paper:
\begin{enumerate}[leftmargin=*,label=(\arabic*)]
    \item We use structural equation models (SEMs) to rigorously define patient-specific root causes as variables associated with the \textit{predictive exogenous error terms} of an SEM (Section \ref{sec_RC}).
    \item We quantify the predictivity of the errors using sample-specific Shapley values  (Section \ref{sec_RC}).
    \item We use the above formulation to derive an algorithm called Root Causal Inference (RCI) that quantifies the ``degree of root cause-ness'' for each variable of each patient by (a) extracting the error terms from the data and then (b) estimating patient-specific Shapley values using these errors (Section \ref{sec_RCI}).
    \item We speed up and customize a critical sub-procedure of RCI to improve scalability in the linear setting (Section \ref{sec_lingam}). 
    \item We show that the proposed algorithm outperforms existing alternatives by uncovering large effect sizes within each patient that are nevertheless inconsequential at the group level (Section \ref{sec_exp}).
\end{enumerate}
\end{tcolorbox}

\section{Structural Equation Models}

We must define a causal process before we can define a root cause. We therefore provide necessary background material on SEMs. 

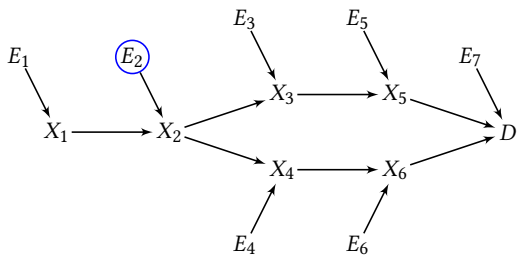
\begin{figure}[b]
\centering
\begin{tikzpicture}[scale=1.0, shorten >=1pt,auto,node distance=2.8cm, semithick]
                    
\tikzset{vertex/.style = {inner sep=0.4pt}}
\tikzset{edge/.style = {->,> = latex'}}
 
\node[vertex] (1) at  (0,0) {$X_1$};
\node[vertex] (2) at  (1.5,0) {$X_2$};
\node[vertex] (3) at  (3,0.5) {$X_3$};
\node[vertex] (4) at  (3,-0.5) {$X_4$};
\node[vertex] (5) at  (4.5,0.5) {$X_5$};
\node[vertex] (6) at  (4.5,-0.5) {$X_6$};
\node[vertex] (7) at  (6,0) {$D$};

\node[vertex] (8) at  (-0.5,1) {$E_1$};
\draw[edge] (8) to (1);
\node[vertex,draw=blue, circle] (9) at  (1,1) {$E_2$};
\draw[edge] (9) to (2);
\node[vertex] (10) at  (2.5,1.5) {$E_3$};
\draw[edge] (10) to (3);
\node[vertex] (11) at  (4,1.5) {$E_5$};
\draw[edge] (11) to (5);
\node[vertex] (12) at  (5.5,1) {$E_7$};
\draw[edge] (12) to (7);
\node[vertex] (13) at  (2.5,-1.5) {$E_4$};
\draw[edge] (13) to (4);
\node[vertex] (14) at  (4,-1.5) {$E_6$};
\draw[edge] (14) to (6);

\draw[edge] (1) to (2);
\draw[edge] (2) to (3);
\draw[edge] (2) to (4);
\draw[edge] (3) to (5);
\draw[edge] (4) to (6);
\draw[edge] (5) to (7);
\draw[edge] (6) to (7);
\end{tikzpicture}

\caption{A different representation of Figure \ref{fig_shock} illustrating the basic idea of using SEMs. We associate each variable with an error term and then attempt to identify the blue error representing the initial shock.} \label{fig_SEM}
\end{figure}

We will use SEMs to construct a directed graph representing a causal process -- like in Figure \ref{fig_shock} -- but with exogenous error terms $\bm{E}$ (Figure \ref{fig_SEM}). Each error term is associated with a variable in $\bm{Z} = \bm{X} \cup D$ and intuitively represents the variable's \textit{unique contribution} to the causal process. The effect of the shock on $X_2$ in Figure \ref{fig_shock} corresponds to the effect of the blue error term on the causal process in Figure \ref{fig_SEM}. The goal is therefore to identify the value of this blue error term for each patient and then quantify its causal effect on $D$. Note that the example only contains one shock, but we can potentially have multiple shocks for each patient.

We now describe the ideas more concretely. An SEM over random variables $\bm{Z}$ corresponds to a series of equations in the form:
\begin{equation} \nonumber
    Z_i = g_i(\textnormal{Pa}(Z_i),E_i), \hspace{3mm} \forall Z_i \in \bm{Z},
\end{equation}
where $\textnormal{Pa}(Z_i)$ refers to the \textnormal{parents}, or direct causes of $Z_i$, and $E_i \in \bm{E}$ to an error term associated with $Z_i$. The random vector $\bm{E}$ corresponds to a set of mutually independent random variables. 

A \textit{directed graph} $\mathbb{G}$ is a graph over the vertices $\bm{Z}$ with at most one directed edge $\rightarrow$ or $\leftarrow$ between any two vertices. We can associate $\mathbb{G}$ with a set of structural equations by drawing a directed edge from $Z_j$ to $Z_i$ if $Z_j \in \textnormal{Pa}(Z_i)$. We can therefore also write $\textnormal{Pa}_{\mathbb{G}}(Z_i)$ to mean the parents of $Z_i$ in the directed graph $\mathbb{G}$. $Z_j$ is an \textit{ancestor} of $Z_i$, denoted by $Z_j \in \textnormal{Anc}_{\mathbb{G}}(Z_i)$, if there exists a directed path in $\mathbb{G}$ from $Z_j$ to $Z_i$ (or $Z_i = Z_j$) so that $Z_j$ is a cause of $Z_i$. Likewise, $Z_j$ is an \textit{descendant} of $Z_i$, if there exists a directed path in $\mathbb{G}$ from $Z_i$ to $Z_j$ (or $Z_i = Z_j$). An \textit{augmented directed graph} $\mathbb{G}^\prime$ is a directed graph over $\bm{Z} \cup \bm{E}$ such that $\textnormal{Pa}_{\mathbb{G}^\prime}(Z_i) = \textnormal{Pa}_{\mathbb{G}}(Z_i) \cup E_i$ for all $Z_i \in \bm{Z}$ and $\textnormal{Pa}_{\mathbb{G}^\prime}(E_i) = \emptyset$ for all $E_i \in \bm{E}$. Figure \ref{fig_shock} is an example of a directed graph, and Figure \ref{fig_SEM} the corresponding augmented directed graph. We provide additional background regarding \textit{d-separation}, \textit{d-connection}, \textit{the global Markov property} and \textit{d-separation faithfulness} in the Supplementary Materials.

Now consider the SEM given by the following set of linear equations:
\begin{equation} \nonumber
\begin{aligned}
    X_1 = \hspace{1mm}& E_1\\
    X_2 = \hspace{1mm}& X_1 \theta_{12} + E_2\\
    X_3 = \hspace{1mm}& X_2  \theta_{23} + E_3,\\
\end{aligned}
\end{equation}
where $\theta$ refers to a matrix of coefficients whose $i^\textnormal{th}$ row and $j^\textnormal{th}$ column is given by $\theta_{ij}$. We can more generally represent a linear structural equation model in matrix format as follows:
\begin{equation} \nonumber
    \bm{X} = \bm{X} \theta + \bm{E}_{1:p},
\end{equation}
where $\mathbb{E}(\bm{X}) = 0$ without loss of generality, and $p = |\bm{X}|$. The exact values of the coefficients $\theta$ are \textit{not} always identifiable from data. However, $\theta$ is identifiable if (1) $\mathbb{G}$ is a acyclic, or a directed acyclic graph (DAG), and (2) the error terms are non-Gaussian \cite{Shimizu06}.\footnote{More specifically, at most one of the error terms in $\bm{E}_{1:p}$ is Gaussian.} The term \textit{Linear Non-Gaussian Acyclic Model} (LiNGAM) refers to this specific setup. We will pay particular attention to LiNGAM in this paper for computational reasons:
\begin{assumptions1} \label{assump_LiNGAM}
$\bm{X}$ follows the LiNGAM model.
\end{assumptions1}

The diagnostic label $D = \bm{Z} \setminus \bm{X}$ specifies whether a subject is a patient with an illness or a healthy control and is therefore binary. We assume that $D$ is a \textit{terminal vertex}, or a vertex without descendants, and linked to $\bm{X}$ via a logistic function:
\begin{assumptions1} \label{assump_terminal}
$D$ is a terminal vertex such that $\mathbb{P}(D|\bm{X}) =\\ \textnormal{logistic}(\bm{X}\beta + \alpha)$ with only $\beta_{\textnormal{Pa}_\mathbb{G}(D)} \not = 0$.
\end{assumptions1}
\noindent This is a reasonable assumption because a scientist who seeks to identify the causes of $D$ will likely use datasets containing measurements of the non-descendants of the diagnosis, such as gene expression levels, clinical laboratory values, imaging or social history. The logistic link also provides a natural extension of the LiNGAM model so that it can handle a binary variable.

\section{Patient-Specific Root Causes as Predictive Exogeneity} \label{sec_RC}

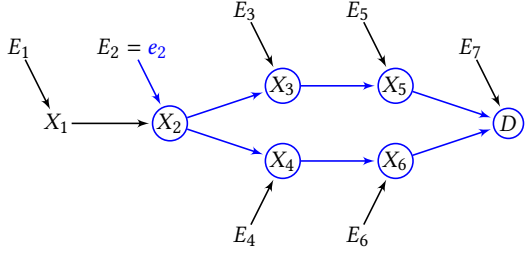
\begin{figure}
\centering
\begin{tikzpicture}[scale=1.0, shorten >=1pt,auto,node distance=2.8cm, semithick]
                    
\tikzset{vertex/.style = {inner sep=0.4pt}}
\tikzset{edge/.style = {->,> = latex'}}
 
\node[vertex] (1) at  (0,0) {$X_1$};
\node[vertex,draw=blue, circle] (2) at  (1.5,0) {$X_2$};
\node[vertex,draw=blue, circle] (3) at  (3,0.5) {$X_3$};
\node[vertex,draw=blue, circle] (4) at  (3,-0.5) {$X_4$};
\node[vertex,draw=blue, circle] (5) at  (4.5,0.5) {$X_5$};
\node[vertex,draw=blue, circle] (6) at  (4.5,-0.5) {$X_6$};
\node[vertex,draw=blue, circle, minimum size=0.40cm] (7) at  (6,0) {$D$};

\node[vertex] (8) at  (-0.5,1) {$E_1$};
\draw[edge] (8) to (1);
\node[vertex] (9) at  (1,1) {$E_2 = \textcolor{blue}{e_2}$};
\draw[edge,blue] (9) to (2);
\node[vertex] (10) at  (2.5,1.5) {$E_3$};
\draw[edge] (10) to (3);
\node[vertex] (11) at  (4,1.5) {$E_5$};
\draw[edge] (11) to (5);
\node[vertex] (13) at  (2.5,-1.5) {$E_4$};
\draw[edge] (13) to (4);
\node[vertex] (14) at  (4,-1.5) {$E_6$};
\draw[edge] (14) to (6);
\node[vertex] (12) at  (5.5,1) {$E_7$};
\draw[edge] (12) to (7);

\draw[edge] (1) to (2);
\draw[edge,blue] (2) to (3);
\draw[edge,blue] (2) to (4);
\draw[edge,blue] (3) to (5);
\draw[edge,blue] (4) to (6);
\draw[edge,blue] (5) to (7);
\draw[edge,blue] (6) to (7);
\end{tikzpicture}

\caption{Perturbing $E_2$ changes all of its descendants, ultimately impacting $D$.} \label{fig_SEM_down}
\end{figure}

As mentioned previously, we model an initial shock to the system as a change in the value of $E_i \in \textnormal{Anc}_{\mathbb{G}^\prime}(D) \setminus E_{p+1}$, where $E_{p+1}$ is the error term associated with $D$. In particular, we can write the following equation for any healthy control:
\begin{equation} \nonumber
    X_i = g_i(\textnormal{Pa}_{\mathbb{G}}(X_i),E_i = \textcolor{blue}{\widetilde{e}_i}),
\end{equation}
where $E_i = \widetilde{e}_i$ corresponds to a ``healthy'' instantiation of $E_i$. Suppose however that there is a shock to the system that changes the normal value $\widetilde{e}_i$ to an ``unhealthy'' one $e_i$: 
\begin{equation} \label{eq_err_change}
    X_i = g_i(\textnormal{Pa}_{\mathbb{G}}(X_i),E_i = \textcolor{blue}{e_i}).
\end{equation}
This change from $\widetilde{e}_i$ to $e_i$ then perturbs the value of $X_i$ and all of its descendants. We represent this cascade down the causal system in Figure \ref{fig_SEM_down}, where a change in the error term highlighted in blue in turn changes the values of all of its descendants.

Not all shocks will ultimately increase the probability of disease, or the probability that $D=1$. But the value of $E_i$ can clearly increase the probability of developing disease so that the following difference of log-odds is much greater than zero:
\begin{equation} \nonumber
    \gamma_{E_i \cup \bm{W}} \triangleq \underbrace{\textnormal{ln} \Big[\frac{\mathbb{P}(D=1| E_i\cup \bm{W})}{\mathbb{P}(D=0|E_i\cup \bm{W})}\Big]}_{(1)}- \underbrace{\textnormal{ln}\Big[\frac{\mathbb{P}(D=1| \bm{W})}{\mathbb{P}(D=0| \bm{W})} \Big]}_{(2)}.
\end{equation} 
If $E_i$ \textit{alone} increases the probability of developing disease, then (1) is larger than (2) when $\bm{W} = \emptyset$, so $\gamma_{E_i}>0$. 

$E_i$ may not have a marginal effect $\gamma_{E_i} > 0$, but it may have a \textit{joint} effect $\gamma_{E_i \cup \bm{W}} > 0$ when present in combination with other errors $\bm{W} \subseteq (\bm{E}_{1:p} \setminus E_i)$. For instance, a single genetic mutation is unlikely to lead to cancerous growth, but multiple mutations often do \cite{Loeb20}. We do not \textit{a priori} know which combination of errors have a joint effect, so we score the contribution of $E_i$ as the sum over all possible combinations of errors that include $E_i$:
\begin{equation} \label{eq_shapley}
    S_i \triangleq \frac{1}{p}\hspace{-15mm}\underbrace{\sum_{\bm{W} \subseteq (\bm{E}_{1:p} \setminus E_i)} \frac{1}{\binom{p-1} {|\bm{W}|}}}_{\textnormal{Average over all possible combinations of } \bm{E}_{1:p} \setminus E_i} \hspace{-14.5mm}\gamma_{E_i \cup \bm{W}}.
\end{equation}
\noindent This is precisely the \textit{sample-specific Shapley value} with log-odds \cite{Strumbelj15,Lundberg17}.

The above derivation leads to the definition of a patient-specific root cause:
\begin{definition1} \label{def_root} (Patient-Specific Root Cause)
     $X_i \in \bm{X}$ is a \textit{patient-specific root cause} if $E_i \in \textnormal{Anc}_{\mathbb{G}^\prime}(D)$ and $S_i > 0$.\footnote{A negative $S_i$ value, on the other hand, implies that the error term is protective and therefore an inducer of health -- not disease.}
\end{definition1}
\noindent In other words, a patient-specific root cause is a variable with an error term that \textit{predictably} induces diseases as quantified by the sample-specific Shapley value in Equation \eqref{eq_shapley}. Notice that a root cause is not necessarily an outlier or even a large shock to the system such that $S_i \gg 0$. Definition \ref{def_root} only requires $S_i > 0$ so that a patient may have multiple root causes, each with a small effect, but \textit{in combination} still culminate into disease.

\section{Detection of Patient-Specific Root Causes} \label{sec_RCI}

We now detail an algorithm to detect patient-specific root causes by capitalizing on Definition \ref{def_root}. In order to identify all patient-specific root causes, we need to recover the exogenous errors that (1) are ancestors of $D$, and (2) have positive sample-specific Shapley values per Definition \ref{def_root}. Satisfying (1) is straightforward when $D$ is a terminal vertex because we must have:
\begin{lemma1} \label{lem_anc}
If $D$ is a terminal vertex, then $D \ci E_i | \bm{W}$ for any $E_i \not \in \textnormal{Anc}_{\mathbb{G}^\prime}(D)$ and $\bm{W} \subseteq (\bm{E} \setminus E_i)$.
\end{lemma1}
\noindent All proofs are located in the Supplementary Materials. The above lemma implies $\mathbb{P}(D|\bm{W} \cup E_i) = \mathbb{P}(D|\bm{W})$ for any $E_i \not \in \textnormal{Anc}_{\mathbb{G}^\prime}(D)$. We thus have $S_i = 0$ in these cases, so we can easily eliminate all $X_i \in \bm{X}$ with $E_i \not \in \textnormal{Anc}_{\mathbb{G}^\prime}(D)$ because any patient-specific root cause must satisfy $S_i > 0$. It therefore suffices to \textit{only} compute $S_i$ for each $X_i \in \bm{X}$ with the terminal vertex assumption.

We now propose an algorithm called Root Causal Inference (RCI) that estimates the sample-specific Shapley value of each variable in $\bm{X}$ for each patient. We summarize RCI in Algorithm \ref{alg_RCI}. The algorithm first runs Direct LiNGAM (DL), an algorithm that estimates the exogenous error terms as $\widehat{\bm{E}}_{1:p}$ \cite{Shimizu11}; we detail the method in Section \ref{sec_lingam}. RCI then estimates $\mathbb{P}(D|\widehat{\bm{E}}_{1:p})$ by logistic regression in Step \ref{alg_RCI:regress}. The Shapley value in Equation \eqref{eq_shapley} averages over an exponential number of terms, but we can compute it efficiently for each $X_i \in \bm{X}$ from the coefficients $\delta$ of logistic regression of $D$ on $\bm{E}_{1:p}$ using the following result:
\begin{proposition1} \label{prop_delta}
$S_i = E_i\delta_i$ under Assumptions \ref{assump_LiNGAM} and \ref{assump_terminal}.
\end{proposition1}
\noindent The proof follows immediately from Corollary 1 in \cite{Lundberg17}. RCI therefore estimates $S_i$ as follows:
\begin{equation} \label{eq_perm}
    \widehat{S}_{i} = \widehat{E}_i\widehat{\delta}_i,
\end{equation}
where $\widehat{\delta}$ corresponds to the coefficients estimated by logistic regression. Let $\widehat{s}_i^k=\widehat{e}_i^k \widehat{\delta}_i$ more specifically refer to the estimated value of $S_i$ for patient $k$, where $\widehat{E}_i = \widehat{e}_i^k$ for patient $k$ in Equation \eqref{eq_perm}. RCI then computes the matrix $\mathcal{S}$ in Step \ref{alg_RCI:delta}, where each row corresponds to a patient in the test set $\mathcal{T}$ and each column to a variable in $\bm{X}$, so $\mathcal{S}_{ki} = \widehat{s}_i^k$. We certify RCI with the following theorem:
\begin{theorem1} \label{thm_RCI}
Under Assumptions \ref{assump_LiNGAM} and \ref{assump_terminal}, RCI outputs the true Shapley values $\mathcal{S}_{k\cdot}$ for any patient $k$ in $\mathcal{T}$ in the large sample limit. Moreover, $s_i^k > 0$ \underline{if and only if} $X_i$ is a root cause of $D$ for patient $k$.
\end{theorem1}

\begin{algorithm}[]
 \nonl \textbf{Input:} $\bm{X}$, test set $\mathcal{T}$\\
 \nonl \textbf{Output:}  $\mathcal{S}$\\
 \BlankLine

$\widehat{\bm{E}}_{1:p} \leftarrow$ Direct LiNGAM($\bm{X}$) \label{alg_RCI:lingam}\\
Estimate $\mathbb{P}(D|\widehat{\bm{E}}_{1:p})$ using logistic regression \label{alg_RCI:regress}\\
Compute the matrix $\mathcal{S}$ containing the estimated sample-specific Shapley values of each patient in $\mathcal{T}$ \label{alg_RCI:delta}

\caption{Root Causal Inference (RCI)} \label{alg_RCI}
\end{algorithm}

\section{Improved Direct LiNGAM} \label{sec_lingam}

The DL algorithm in Line \ref{alg_RCI:lingam} of RCI unfortunately has some shortcomings out of box:
\begin{enumerate}[leftmargin=*,label=(\arabic*)]
    \item The algorithm is slow; it scales $O(p^3)$ and therefore struggles when datasets contains hundreds of variables.
    \item The algorithm is sample inefficient; it recovers the error terms of \textit{all} of the variables in $\bm{X}$, even though we are only interested in the errors of the ancestors of $D$ per Definition \ref{def_root}.
\end{enumerate}
We rectify both of these problems by changing DL in several respects.

\subsection{Faster Score Minimization}
We first require a deep understanding of the DL algorithm in order to address (1). We summarize the parts of DL applicable to this paper in Algorithm \ref{alg_DL}. The procedure starts with the set $\bm{U} = [p] \triangleq \{1,\dots,p\}$ in Line \ref{alg_DL:start}. The algorithm then uses the function FindRoot in Line \ref{alg_DL:root} to find an exogenous variable $G$ that is removed from $\bm{U}$ (Line \ref{alg_DL:remove}) and appended to $\bm{K}$ (Line \ref{alg_DL:add}) in each iteration. DL also partials out $G$ from $\bm{X}$ in Line \ref{alg_DL:partial} in each iteration; this step allows DL to extract the mutually independent errors $\bm{E}_{1:p}$ using a linear combination of $\bm{X}$ similar to independent component analysis (ICA) \cite{Hyvarinen13}.

FindRoot is the most computationally intensive part of DL. Let $R_{ij}$ denote the residuals of $X_i$ when linearly regressed on $X_j$. We have the following well-known result:
\begin{proposition1} (Lemma 1 in \cite{Shimizu11}) \label{prop_ind}
Under the LiNGAM model, $X_j$ is an exogenous variable if and only if $X_j \ci R_{ij}$ for all $X_i \in \bm{X} \setminus X_j$.
\end{proposition1}
\noindent FindRoot therefore identifies an exogenous variable in each iteration by finding the variable most independent of its residuals.

We detail FindRoot in Algorithm \ref{alg_root}. The original algorithm in \cite{Shimizu11} performs all pairwise comparisons of the entries in $\bm{U}$, or $|\bm{U}|(|\bm{U}|-1)$ comparisons. The algorithm keeps track of a score quantifying the degree of dependence for each variable in $\bm{U}$ within the vector $\bm{T}$ initialized in Line \ref{alg_root:S}. The score is a sum of quantities like $\textnormal{min}(0,\mathcal{C}_{ij})^2$. We set $\mathcal{C}_{ij}$ to the measure of independence proposed in \cite{Hyvarinen13}:
\begin{equation} \nonumber
\begin{aligned}
    \mathcal{C}_{ij} &= -H(X_i) - H(R_{ji}) + H(X_j) + H(R_{ij}),
\end{aligned}
\end{equation}
 once all random variables are standardized. $\mathcal{C}_{ij}$ is greater than zero when $X_j \ci R_{ij}$. The function $H$ denotes an approximate version of differential entropy given by:
\begin{equation} \nonumber
\begin{aligned}
H(Y) = (1+\textnormal{log} 2 \pi)/2 - &k_1[\mathbb{E} (\textnormal{log cosh}(Y)) - \delta]^2\\ - &k_2[\mathbb{E}(Y\textnormal{exp}(-Y^2/2)]^2,
\end{aligned}
\end{equation}
with $k_1=79.047, k_2 = 7.4129, \delta = 0.37457$.
Observe that $\mathcal{C}_{ij} = -\mathcal{C}_{ji}$; this property allows us to reduce the number of pairwise comparisons from $|\bm{U}|(|\bm{U}|-1)$ to $|\bm{U}|(|\bm{U}|-1)/2$ in Lines \ref{alg_root:loop1} and \ref{alg_root:loop2} of Algorithm \ref{alg_DL}. Many other measures of independence exist, but we choose this measure due to its computational efficiency and accuracy. FindRoot finally outputs the variable in $\bm{U}$ associated with the minimum score in Line \ref{alg_root:min}; in other words, the algorithm outputs the most independent exogenous variable in accordance with Proposition \ref{prop_ind}.

FindRoot unfortunately \textit{always} performs $|\bm{U}|(|\bm{U}|-1)/2$ iterations -- regardless of the problem at hand. We may however only need to perform a small subset of these iterations for a given problem. We therefore propose FindRoot+ which performs just enough comparisons to find the variable in $\bm{U}$ associated with the minimum score. 

 We summarize FindRoot+ in Algorithm \ref{alg_rootp}. The algorithm only needs to find the variable associated with the minimum score in $\bm{T}$ after all iterations, so FindRoot+ checks whether the minimum score in $\bm{T}$ remains the smallest at each iteration rather than updating all scores in $\bm{T}$ like FindRoot. In particular, FindRoot+ first finds the indices $I$ already associated with the smallest score in $\bm{T}$ in Line \ref{alg_rootp:I}. The algorithm then only updates those indices in $\bm{T}$ in Lines \ref{alg_rootp:compare1} and \ref{alg_rootp:compare2}. The vector $\bm{O}$ keeps track of which variables to regress on next, and the matrix $M$ keeps track of which regressions have been performed to avoid redundant computations. FindRoot+ breaks from the repeat loop in Lines \ref{alg_rootp:O}-\ref{alg_rootp:Obreak} once the algorithm finds a variable with (a) the minimum score in $\bm{T}$ and (b) all necessary regressions completed according to $\bm{O}$; this ensures that the algorithm stops as soon as possible. The algorithm therefore ultimately outputs the variable in $\bm{U}$ associated with the minimum score in Line \ref{alg_rootp:min} just like FindRoot. However, FindRoot+ also adapts to the problem at hand by only performing just enough computations to find the minimum score. We provide the following guarantee:
 \begin{proposition1} \label{prop_root}
 FindRoot+ and FindRoot always output the same root variable $G$.
 \end{proposition1}
 \noindent Experiments in Section \ref{sec_exp} show that FindRoot+ speeds up DL by about 1.5-3 times on average when substituted in place of FindRoot. 

\begin{algorithm}[t]
 \nonl \textbf{Input:} $\bm{X}$\\
 \nonl \textbf{Output:}  $\bm{E}_{\bm{K}},\bm{K}$\\
 \BlankLine

$\bm{U} \leftarrow [p]$\\ \label{alg_DL:start}
$\bm{K} \leftarrow \emptyset$\\
\Repeat{$\bm{U} = \emptyset$}{
      $G \leftarrow$ FindRoot($\bm{X},\bm{U}$)\\ \label{alg_DL:root}
        $\bm{U} \leftarrow \bm{U} \setminus G$\\ \label{alg_DL:remove}
        $\bm{K} \leftarrow \bm{K} \cup G$\\ \label{alg_DL:add}
      $\bm{X} \leftarrow $ partial out $G$ from $\bm{X}$ \label{alg_DL:partial}
    }
$\bm{E}_{\bm{K}} \leftarrow \bm{X}_{\bm{K}}$ \label{alg_DL:errors}
\caption{Direct LiNGAM} \label{alg_DL}
\end{algorithm}

\begin{algorithm}[t]
 \nonl \textbf{Input:} $\bm{X}, \bm{U}$\\
 \nonl \textbf{Output:} root $G$\\
 \BlankLine

\textbf{return} $\bm{U}$  if $|\bm{U}| = 1$\\
$\bm{T} = \bm{0}_{|\bm{U}|}$ \label{alg_root:S}\\
\For{$i \in [|\bm{U}|-1]$ \label{alg_root:loop1}}{
\For{$j \in \{i+1, \dots, |\bm{U}|\}$ \label{alg_root:loop2}}{
      $T_i = T_i + \textnormal{min}(0,\mathcal{C}_{ij})^2$\\
      $T_j = T_j + \textnormal{min}(0,-\mathcal{C}_{ij})^2$
    }
}
$G \leftarrow \bm{U}[\argmin_i T_i]$ \label{alg_root:min}

\caption{FindRoot} \label{alg_root}
\end{algorithm}

\begin{algorithm}[t]
 \nonl \textbf{Input:} $\bm{X}, \bm{U}$\\
 \nonl \textbf{Output:} root $G$\\
 \BlankLine

\textbf{return} $\bm{U}$ if $|\bm{U}| = 1$\\
$\bm{T} = \bm{0}_{|\bm{U}|}$\\
$\bm{O} = \bm{1}_{|\bm{U}|}$\\
$M = \textnormal{TRUE}_{|\bm{U}| \times |\bm{U}|}$\\
\SetKwBlock{Repeat}{repeat}{end}
\Repeat{
    $I \leftarrow $ indices associated with minimum value in $\bm{T}$\\ \label{alg_rootp:I}
    \If{$\exists O_i \in \bm{O}$ such that $O_i> |\bm{U}|$ and $i \in I$ \label{alg_rootp:O}}{\textbf{break}} \label{alg_rootp:Obreak}
    \For{$i \in I$}{
        \If{$M[i,O_i]$}{
        $T_i \leftarrow T_i + \textnormal{min}(0, \mathcal{C}_{iO_i})^2$\\ \label{alg_rootp:compare1}
        $T_{O_i} \leftarrow T_{O_i} + \textnormal{min}(0, -\mathcal{C}_{iO_i})^2$\\ \label{alg_rootp:compare2}
        $M[i,O_i] = $ FALSE\\
        $M[O_i,i] = $ FALSE\\
        }
        $O_i = O_i + 1$\\ \label{alg_rootp:increment}
    }
}
$G \leftarrow \bm{U}[\argmin_i T_i]$ \label{alg_rootp:min}
\caption{FindRoot+} \label{alg_rootp}
\end{algorithm}

\subsection{Localized Partial Order}

We now address (2) and further improve on (1). \textcolor{blue}{Plus} DL, or DL equipped with \textcolor{blue}{FindRoot+}, recovers the error terms associated with all of the variables in $\bm{X}$. RCI however only requires the error terms in $\textnormal{Anc}_{\mathbb{G}^\prime}(D) \setminus E_{p+1}$. We therefore also propose Local Plus DL which only recovers the error terms in $\textnormal{Anc}_{\mathbb{G}^\prime}(D)  \setminus E_{p+1}$ using the FindRoot+ algorithm. 

We summarize Local Plus in Algorithm \ref{alg_LDL}. The algorithm differs from the original version of DL and Plus DL at Lines \ref{alg_LDL:ancDs}-\ref{alg_LDL:ancDe}; we use the notation $p \leftarrow X_i \ci D$ to mean that the independence oracle returns $1$ if $X_i \ci D$ and $0$ otherwise. Lines \ref{alg_LDL:ancDs}-\ref{alg_LDL:ancDe} determine whether $X_i \ci D$ for all $X_i \in \bm{U}$ due to the following result:
\begin{lemma1} \label{lem_error}
Consider any $E_i \in \bm{E} \setminus E_{p+1}$. Under d-separation faithfulness and Assumption \ref{assump_terminal}, $E_i \in \textnormal{Anc}_{\mathbb{G}^\prime}(D)$ \underline{if and only if} $E_i \not \ci D$.
\end{lemma1}
\noindent D-separation faithfulness is a mild condition defined in the Supplementary Materials that enables a one-to-one correspondence between d-separation and conditional independence. Lines \ref{alg_LDL:ancDs}-\ref{alg_LDL:ancDe} therefore eliminate any error term $E_i$ such that $E_i \not \in \textnormal{Anc}_{\mathbb{G}^\prime}(D)$ from $\bm{U}$. Local Plus then iteratively identifies an error term in Line \ref{alg_LDL:rootp} using FindRoot+ and then partials out that error term in Line \ref{alg_LDL:partial}. Repeating this process until $\bm{U}$ is empty ensures that the algorithm identifies all and only the error terms in $\textnormal{Anc}_{\mathbb{G}^\prime}(D) \setminus E_{p+1}$:
\begin{theorem1} \label{thm_LP}
Under d-separation faithfulness and Assumptions \ref{assump_LiNGAM} and \ref{assump_terminal}, Local Plus \underline{exclusively} recovers the error terms in $\textnormal{Anc}_{\mathbb{G}^\prime}(D) \setminus E_{p+1}$ for any $0 < \alpha \leq 1$.
\end{theorem1}

\textbf{Time Complexity.} Finally observe that Lines \ref{alg_LDL:ancDs}-\ref{alg_LDL:ancDe} reduce the complexity of DL from $O(p^3)$ to $O(r^3)$, where $r$ denotes the number of variables d-connected to $D$ and typically $r \ll p$. Experiments in Section \ref{sec_exp:Synth} show that Local Plus is around 4-250 times faster than the original version of DL, depending on the size of $p$, even with a liberal $\alpha$ threshold of 0.2.
\begin{algorithm}[t]
  \nonl \textbf{Input:} $\bm{X}, D$, $\alpha$\\
 \nonl \textbf{Output:}  $\bm{E}_{\bm{K}},\bm{K}$\\
 \BlankLine

$\bm{U} \leftarrow \bm{X}$\\
$\bm{K} \leftarrow \emptyset$\\
\Repeat{$U = \emptyset$}{
      \For{$X_i \in \bm{U}$  \label{alg_LDL:ancDs} \tikzmark{top}}{
$p \leftarrow X_i \ci D$\\
\If{ $p \geq \alpha$}{
    $\bm{U} \leftarrow \bm{U} \setminus X_i$ \tikzmark{right}
    }} \label{alg_LDL:ancDe}
      $G \leftarrow$ FindRoot+($\bm{X},\bm{U}$)\\ \label{alg_LDL:rootp}
      $\bm{U} \leftarrow \bm{U} \setminus G$\\
      $\bm{K} \leftarrow \bm{K} \cup G$\\
      $\bm{X} \leftarrow $ partial out $G$ from $\bm{X}$\\ \label{alg_LDL:partial}
    }
    $\bm{E}_{\bm{K}} \leftarrow \bm{X}_{\bm{K}}$
    \AddNote{top}{}{right}{Removes non-ancestors of $D$ from $\bm{U}$}
\caption{Local Plus Direct LiNGAM} \label{alg_LDL}
\end{algorithm}

\section{Related Work} \label{sec_RW}

Local Plus and RCI were both inspired by the work in \cite{Lasko19}. The authors performed ICA on electronic health record data collected from patients with liver disease; ICA extracts mutually independent error terms just like Local Plus in Step \ref{alg_RCI:lingam} of RCI. The authors then used the error terms to predict the presence of hepatocellular carcinoma 10 years later using random forest; the binary variable $D$ thus denotes the absence or presence of hepatocellular carcinoma and corresponds to a \textit{terminal vertex} in this case per Assumption \ref{assump_terminal}. The most important error terms, as assessed by a random forest permutation measure, exactly matched \textit{in ranked order} the top causes of liver transplantation. This suggested that the predictive accuracy of exogenous errors accurately identifies the largest ``shocks'' that cause hepatocellular carcinoma. However, the authors could not explain \textit{why} extracting mutually independent features yielded such interesting results. They also did not identify \textit{patient-specific} root causes.

The second most closely related paper is \cite{Janzing19}, where the authors defined a patient-specific root cause $X_i$ as an outlier according to the conditional distribution $\mathbb{P}(X_i | \textnormal{Pa}_{\mathbb{G}}(X_i))$. An outlier however does not always increase the probability of developing a disease, and not all diseases are caused by outliers. We therefore instead define root causes as variables with predictive error terms that are ancestors of $D$ per Definition \ref{def_root}. Note that the authors proposed another variation in \cite{Budhathoki21}, focusing instead on identifying the root causes using Shapley values computed by substituting causal conditionals. This algorithm however modifies the SEM by substituting certain conditional probabilities in the factorization of the DAG. As a result, it cannot be used to compute patient-specific root causes, even though we really need to identify root causes \textit{at the subject level} in order to make complex diseases tractable. The method also does not scale to more than several variables, since the number of substitutions required to compute the Shapley values grows exponentially with the number of variables.

Several other approaches exist for inferring changes in SEMs, but they all equate the diagnosis to a context that \textit{perfectly} indexes disease versus healthy. Examples include difference of DAGs \cite{Wang18,Ghoshal19} and the discovery of graphs summarizing multiple contexts \cite{Huang20,Mooij20}. The diagnosis $D$ is however not a context, but a corrupted estimate of the context inferred from clinical judgment based on downstream signs and symptoms. The context formulation also implies that the diagnosis is a parent of the root causes, which does not make sense because the diagnosis must be an effect of the root causes. These algorithms are therefore more suitable for discovering causal relationships under \textit{perfectly known} shifts in the environment, such as in experiments or with different medications.

Note that many other researchers have attempted to uncover latent factors by using other principles not necessarily related to causal inference. For example, sparse non-negative tensor factorization decomposes data into a sum of low rank non-negative tensors \cite{Ho14,Zhou14}. The correspondence between these factorizations and causation remains unclear. A variety of clustering methods inherently assume that each patient falls into a \textit{single discrete} cluster when, in reality, the illness of each patient may vary due to multiple factors that do not necessarily form distinct regions of high density \cite{Marlin12,Schulam15}. Finally, many deep learning methods extract latent representations which, while predictive, do not necessarily encode causal relationships \cite{Lasko13,Kale14,Che15}. One algorithm deduces the causal relationship between the \textit{latent representation} of a neural network and $D$ \cite{Kale15}. However, we actually want to discover the causal relationships between the \textit{original variables} $\bm{X}$ and $D$ because treatments ultimately target real-world entities $\bm{X}$ rather than the representations of a neural network. We conclude that none of the above tensor factorization, clustering or multi-layer representation methods attempt to identify patient-specific root causes in a principled manner.

\section{Comparison to Classical Approaches}

We cannot finish this paper without clearly discussing the difference between the proposed ideas and traditional methods of differential hypothesis testing or predictive accuracy. We have the following definition for the former:
\begin{definition1} \label{def_diff} (Differential Variable)
    A random variable $X_i$ is \textit{differential} if $X_i \not \ci D$ -- i.e., $\mathbb{P}(X_i|D) \not = \mathbb{P}(X_i)$.
\end{definition1}
\noindent Standard t-tests determine whether $\mathbb{P}(X_i|D) = \mathbb{P}(X_i)$ by specifically testing the likelihood that $\mathbb{E}(X_i|D) = \mathbb{E}(X_i)$. In contrast, F-tests focus on the equality of the variances. Notice that a differential variable differs from a root cause because a differential variable is not necessarily an ancestor of $D$ or related to the predictivity of its exogenous error. A differential variable simply corresponds to the set of variables unconditionally d-connected to $D$ in a DAG model \cite{Spirtes00,Pearl98}.

We next have:
\begin{definition1} \label{def_pred} (Predictive Variable)
    A random variable $X_i$ is \textit{predictive} if $D \not \ci X_i | (\bm{X} \setminus X_i)$ -- i.e., $\mathbb{P}(D|\bm{X} \setminus X_i) \not = \mathbb{P}(D|\bm{X})$.
\end{definition1}
\noindent This again differs from the definition of a root cause for the same reasons: a predictive variable is not necessarily an ancestor of $D$ or related to the predictivity of its error. The predictive variables in fact comprise the \textit{Markov boundary}, or the direct causes, direct effects and direct causes of the direct effects of $D$ in a DAG model \cite{Pearl98,Statnikov13}. The predictive variables therefore simply correspond to the variables nearby $D$ in $\mathbb{G}$ rather than to the root causes, which may lie far from $D$.

We conclude that standard statistical notions of differential distributions or predictive accuracy do not coincide with the proposed definition of patient-specific root causes. Thus, nearly all hypothesis tests or predictive algorithms cannot identify patient-specific root causes -- no matter how powerful the method.

\section{Experiments} \label{sec_exp}

We compare (1) RCI equipped with Local Plus against the following four algorithms representing the state of the art as discussed in the Related Work section:
\begin{enumerate}[leftmargin=*,label=(\arabic*)] \addtocounter{enumi}{1}
    \item Prediction with ICA (ICA): runs ICA and then ranks the identified sources using the local variable importance measure of random forest \cite{Lasko19}. We use the Hungarian algorithm to assign each source to a variable in $\bm{X}$, as described in Step 2 of Algorithm A in \cite{Shimizu11};
    \item Conditional Outliers (CO): learns the causal graph $\widehat{\mathbb{G}}$ with DL and then ranks the ancestors of $D$ in descending order for each patient according to the outlier score $\frac{|X_i - \mathbb{E}(X_i|\textnormal{Pa}_{\widehat{\mathbb{G}}}(X_i))|}{\sigma(X_i|\textnormal{Pa}_{\widehat{\mathbb{G}}}(X_i))}$ \cite{Janzing19};
    \item Model substitution (MS): learns a causal graph with DL, extracts the error terms by diagnostic category, and then ranks the original variables according to Shapley values with mean log-odds using the author proposed substitution of causal conditionals. The method does not scale to many variables, but we can compute the marginal contributions of each variable \cite{Budhathoki21}. 
    \item Semi-interleaved Hiton-PC (HPC): learns the direct causes and direct effects of $D$, denoted by $\bm{P}$, and then ranks the variables using logistic regression according to $\bm{P} \widehat{\beta}$ \cite{Aliferis10}. This strategy is a targeted instantiation of the joint causal inference framework \cite{Mooij20}. We set the $\alpha$ hyperparameter to the default value $0.01$.
\end{enumerate}
We also compared RCI against the following two baselines that, while simple, are commonly used in practice:
\begin{enumerate}[leftmargin=*,label=(\arabic*)] \addtocounter{enumi}{5}
    \item T-tests (TT): ranks variables according to the magnitude of their t-statistics;
    \item Logistic regression with L1 penalty: ranks variables according $\bm{X} \widehat{\beta}$. We use an adaptive penalty to ensure variable selection consistency \cite{Zou06}.
\end{enumerate}
We make sure to standardize the data in order to prevent any gaming of the marginal variances \cite{Reisach21}.
\\${}$\\
\noindent \textbf{Reproducibility.} We make code publicly available, including all algorithms and experimental setups.

\subsection{Synthetic Data} \label{sec_exp:Synth} 

\begin{table*}[t]
\begin{subtable}{0.45\textwidth}  
\centering
\captionsetup{justification=centering,margin=2cm}
\begin{tabular}{cc|ccccccc}
\hhline{=========}
\textit{n} & \textit{p} & RCI            & ICA   & CO    & MS    & HPC            & TT             & LR    \\ \hline
100      & 10         & \textbf{0.755} & 0.416 & 0.439 & 0.369 & 0.591          & 0.588          & 0.585 \\
1,000     & 10         & \textbf{0.927} & 0.506 & 0.471 & 0.374 & 0.705          & 0.625          & 0.642 \\
10,000     & 10         & \textbf{0.976} & 0.597 & 0.468 & 0.408 & 0.726          & 0.610          & 0.654 \\ \hline
100      & 50         & \textbf{0.528} & 0.250 & 0.249 & 0.280 & \textbf{0.492}          & \textbf{0.490}          & 0.394 \\
1,000     & 50         & \textbf{0.840} & 0.488 & 0.330 & 0.282 & 0.665          & 0.586          & 0.584 \\
10,000     & 50         & \textbf{0.956} & 0.523 & 0.297 & 0.272 & 0.649          & 0.552          & 0.552 \\ \hline
100      & 100        & \textbf{0.472} & 0.166 & 0.248 & 0.258 & \textbf{0.498} & \textbf{0.490} & 0.418 \\
1,000     & 100        & \textbf{0.798} & 0.274 & 0.282 & 0.244 & 0.682          & 0.581          & 0.572 \\
10,000     & 100        & \textbf{0.945} & 0.565 & 0.314 & 0.309 & 0.707          & 0.589          & 0.580 \\
\hhline{=========}
\end{tabular}
\caption{RBO} \label{exp_synth:RBO}
\end{subtable} \hspace{15mm} \begin{subtable}{0.45\textwidth}  
\centering
\captionsetup{justification=centering,margin=2cm}
\begin{tabular}{cc|cccc}
\hhline{======}
\textit{n} & \textit{p} & RCI              & MS      & HPC              & LR      \\  \hline
100        & 10         & \textbf{1.06E\Minus1} & 1.32E\Plus0 & \textbf{3.23E\Minus1} & 4.16E\Minus1 \\
1,000      & 10         & \textbf{9.64E\Minus3} & 4.06E\Minus1 & 1.71E\Minus1          & 4.40E\Plus0 \\
10,000     & 10         & \textbf{1.89E\Minus3} & 2.41E\Minus1 & 2.03E\Minus1          & 4.55E\Minus1 \\ \hline
100        & 50         & \textbf{5.33E\Minus2} & 1.01E\Plus1 & \textbf{5.87E\Minus2} & 8.10E\Minus2 \\
1,000      & 50         & \textbf{4.41E\Minus3} & 8.14E\Minus2 & 4.48E\Minus2          & 7.88E\Minus2 \\
10,000     & 50         & \textbf{5.27E\Minus4} & 4.41E\Minus2 & 5.17E\Minus2          & 4.28E\Minus1 \\ \hline
100        & 100        & 8.57E\Minus2          & 4.73E\Plus0 & \textbf{3.08E\Minus2} & 4.54E\Minus2 \\
1,000      & 100        & \textbf{3.67E\Minus3} & 3.34E\Minus2 & 1.36E\Minus2          & 1.13E\Minus1 \\
10,000     & 100        & \textbf{3.67E\Minus4} & 3.57E\Minus2 & 1.92E\Minus2         & 3.12E\Minus1\\
\hhline{======}
\end{tabular}
\caption{MSE}  \label{exp_synth:L2}
\end{subtable}
\vspace{5mm}
\begin{subtable}{0.35\textwidth}  
\centering
\captionsetup{justification=centering,margin=2cm}
\begin{tabular}{cc|cccc}
\hhline{======}
\textit{n} & \textit{p} & Local Plus      & Local  & Plus & Original \\ \hline
100        & 10         & \textbf{4.51}   & 3.21   & 1.04 & 1        \\
1,000      & 10         & \textbf{6.13}   & 4.55   & 1.37 & 1        \\
10,000     & 10         & \textbf{4.62}   & 3.42   & 1.50 & 1        \\ \hline
100        & 50         & \textbf{109.22} & 68.21  & 1.86 & 1        \\
1,000      & 50         & \textbf{61.42}  & 33.83  & 2.06 & 1        \\
10,000     & 50         & \textbf{56.83}  & 27.96  & 2.08 & 1        \\ \hline
100        & 100        & \textbf{261.00} & 156.60 & 1.90 & 1        \\
1,000      & 100        & \textbf{160.09} & 84.53  & 2.00 & 1        \\
10,000     & 100        & \textbf{120.68} & 58.26  & 2.10 & 1       \\
\hhline{======}
\end{tabular}
\caption{Speed-up}  \label{exp_synth:time}
\end{subtable}
\begin{subfigure}{0.635\textwidth}  
\centering
\hspace*{8mm}\includegraphics[scale=0.5]{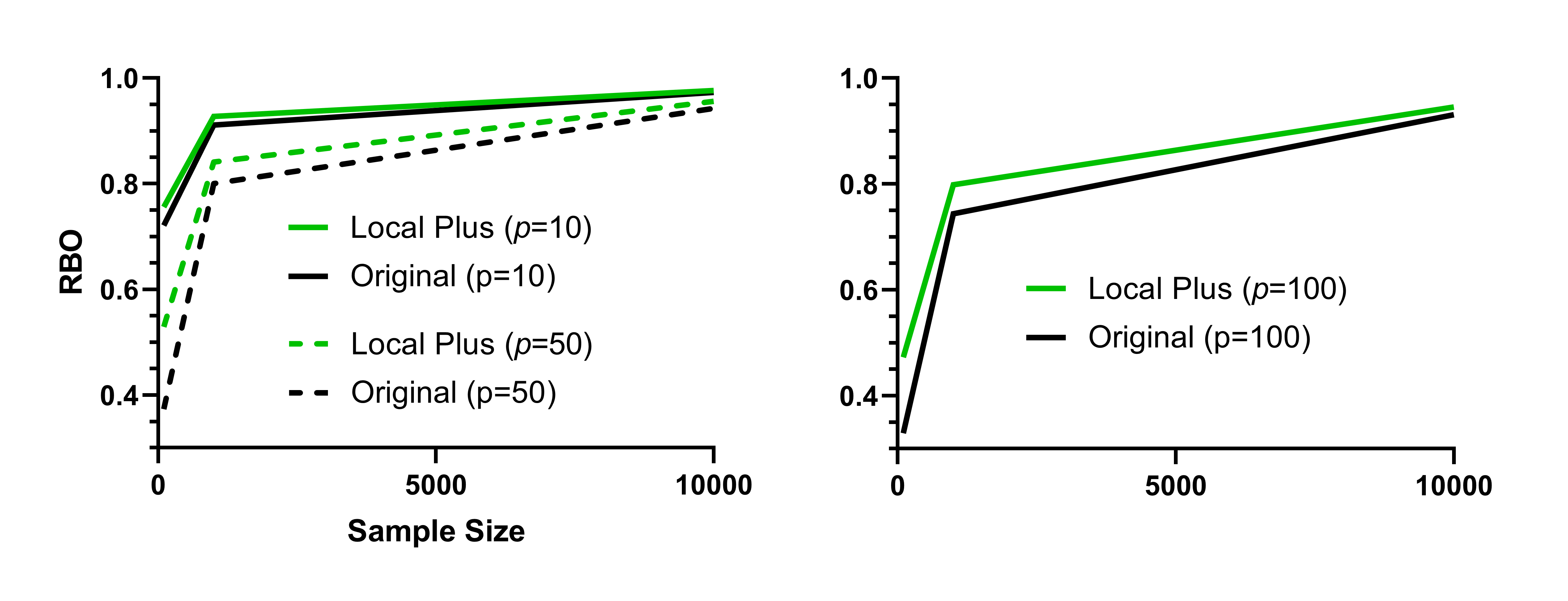}
\caption{}  \label{exp_synth:timing_RBO}
\end{subfigure}
\caption{Synthetic data results in terms of (a) RBO, where higher is better, (b) MSE, where lower is better, and (c) relative speed-up, where higher is better. Bolded values highlight the best performance per row. RCI performs the best in nearly all cases for (a) and (b), often by a very large margin, and continues to improve with increasing sample size. Local Plus speeds up RCI by multiple factors in (c) and is even slightly more accurate than the original version of DL in (d). } \label{exp_synth}
\end{table*}

\subsubsection{Data Generation}
We generated a linear structural equation model with an expected neighborhood size of two, denoted by $\mathbb{E}(N) = 2$, using the following procedure. First, we created a random adjacency matrix $\theta$ with independent realizations of a $\textnormal{Bernoulli}(\mathbb{E}(N)/(p-1))$ variable in the upper triangle of the matrix and zeros otherwise. We then replaced the ones in the matrix by independent realizations of a $\textnormal{Uniform}([-1,-0.25]\cup[0.25,1])$ variable. We instantiated the error term for each variable by sampling from one of the following non-Gaussian distributions chosen uniformly at random: the t-distribution with five degrees of freedom, the chi-square distribution with three degrees of freedom, and the uniform distribution from -1 to 1. We sampled $D$ according to a Bernoulli random variable with probabilities defined by a logistic link function per Assumption \ref{assump_terminal}. We repeated the above procedure 100 times for sample sizes of 100, 1000 and 10000 and dimensions of 10, 50 and 100. We therefore generated a total of $100 \times 3 \times 3 = 900$ independent datasets each obeying a randomly generated model satisfying Assumptions \ref{assump_LiNGAM} and \ref{assump_terminal}.

\subsubsection{Evaluation Criteria}
We evaluated the algorithms using two criteria. Not all algorithms output Shapley values, but each algorithm can output a ranked list of variables. The top ranked variables should ideally correspond to the true root causes with largest effect for each patient. We therefore first compared $\mathcal{R}^k$, the ground truth ranking obtained using the true log-odds Shapley values of all of the root causes for each patient $k$, to the estimated ranking $\widehat{\mathcal{R}}^k$ as given by each algorithm. The notation $\mathcal{R}_{1:i}^k$ refers to the first $i$ variables in the ranking $\mathcal{R}^k$. We utilized the following rank-biased overlap (RBO) measure proposed in Equation (5) of \cite{Webber10}, an established metric that (a) compares ranked lists of potentially varying lengths, (b) weighs top ranks more heavily, and (c) increases monotonically with depth:
\begin{equation} \label{eq:RBO_shapley}
    \frac{1}{n} \sum_{k=1}^n \sum_{i=1}^{q_k} \widetilde{s}_i^k | \widehat{\mathcal{R}}_{1:i}^k \cap \mathcal{R}_{1:i}^k|/i,
\end{equation}
where $s_i^k$ denotes the true Shapley value of $X_i$ for patient $k$, $\widetilde{s}_i^k = \frac{s_i^k}{\sum_{i=1}^{q_k} s_i^k}$ the version normalized to sum to one, and $q_k$ the total number of root causes for patient $k$. The normalized Shapley values and the division by $i$ both ensure that RBO is also normalized to the interval $[0,1]$ for each patient. RBO equals one when $\mathcal{R}^k$ and the top variables of $\widehat{\mathcal{R}}^k$ coincide. On the other hand, RBO equals zero when there is no overlap. A higher value of the metric therefore corresponds to better performance.

RCI, MS, HPC and LR all compute log-odds Shapley values or similar measures. We therefore compare the estimates of these algorithms to the ground truth Shapley values using the mean squared error (MSE):
\begin{equation} \nonumber
    \frac{1}{np} \sum_{k=1}^n \sum_{i=1}^p (\widehat{s}_i^k - s_i^k)^2,
\end{equation}
where lower is better.

\subsubsection{Results}

We report the mean RBO values in Table \ref{exp_synth:RBO}. Bolded values denote the best performance according to paired t-tests at the Bonferonni corrected threshold of 0.05/6, since we compare RCI against 6 other algorithms in each row. RCI obtained the highest mean RBO values compared to all other algorithms across all sample sizes and dimensions. The algorithm tied HPC and TT only in the lowest sample size and higher dimensional settings of $p=50$ and $p=100$. Note that RCI continued to improve with increasing sample sizes, whereas all other algorithms tapered before reaching perfect accuracy. RBO therefore outperformed all algorithms by at least 30\% with sample size $n=10,000$.

The mean MSE values in Table \ref{exp_synth:L2} show similar behavior. RCI again performed the best in nearly all cases. RCI and HPC performed similarly when $n=100$, but RCI quickly overtook HPC once the sample size grew larger. RCI eventually outperformed HPC by one to two orders of magnitude when $n=10,000$.

We display the timing results of an ablation study with four algorithms: Local Plus, Local (or DL with FindRoot and Lines \ref{alg_LDL:ancDs}-\ref{alg_LDL:ancDe} of Algorithm \ref{alg_LDL}),  Plus, and the original version of DL in Table \ref{exp_synth:time}. Local Plus sped up the original version by at least a factor of 4. The speed-up grew with increasing dimensions. At $p=100$, Local Plus was already at least two orders of magnitude faster than the original version, even with a sample size of only 100. We report timing results for Local Plus scaled up to $p=1000$ variables in Table \ref{table_sec} of the Supplementary Materials, as promised in point (1) of Section \ref{sec_lingam}. Finally, Local Plus was actually slightly more accurate than the original version across all dimension sizes (inset in Table \ref{exp_synth:timing_RBO}).

We conclude that RCI is more accurate than all existing alternatives, whether we assess performance by RBO or MSE. Local Plus also speeds up RCI by multiple factors as compared to the original version of DL while improving accuracy.

\subsection{Real Data}

We can identify patient-specific root causes with real data using prior knowledge either in the form of clinical facts or time. We use the following RBO metric to quantify performance similar to Expression \eqref{eq:RBO_shapley}:
\begin{equation} \nonumber
    \frac{1}{n} \sum_{k=1}^n \sum_{i=1}^{q_k} \frac{1}{q_k} | \widehat{\mathcal{R}}_{1:i}^k \cap \mathcal{R}_{1:i}^k|/i,
\end{equation}
where we no longer weight the score by Shapley values, since we do not have access to the true Shapley values with real data.

\subsubsection{Mayo Clinic Primary Biliary Cholangitis}

We first assess performance using clinical knowledge. The Mayo Clinic Primary Biliary Cholangitis (PBC) dataset contains samples of 258 patients with PBC as measured in a randomized clinical trial assessing the effects of a medication called D-penicillamine \cite{Fleming11}. PBC is an autoimmune disease that causes destruction of the small bile ducts in the liver \cite{Hirschfield13}. The disease slowly progresses, eventually causing liver cirrhosis, liver decompensation and then death. The dataset contains several continuous variables including: age, bilirubin, albumin, alkaline phosphatase, copper, cholesterol, platelets, AST and pro-thrombin time. 

\begin{figure*}[t]
\begin{subfigure}{0.23\textwidth}
\centering
\includegraphics[scale=0.5]{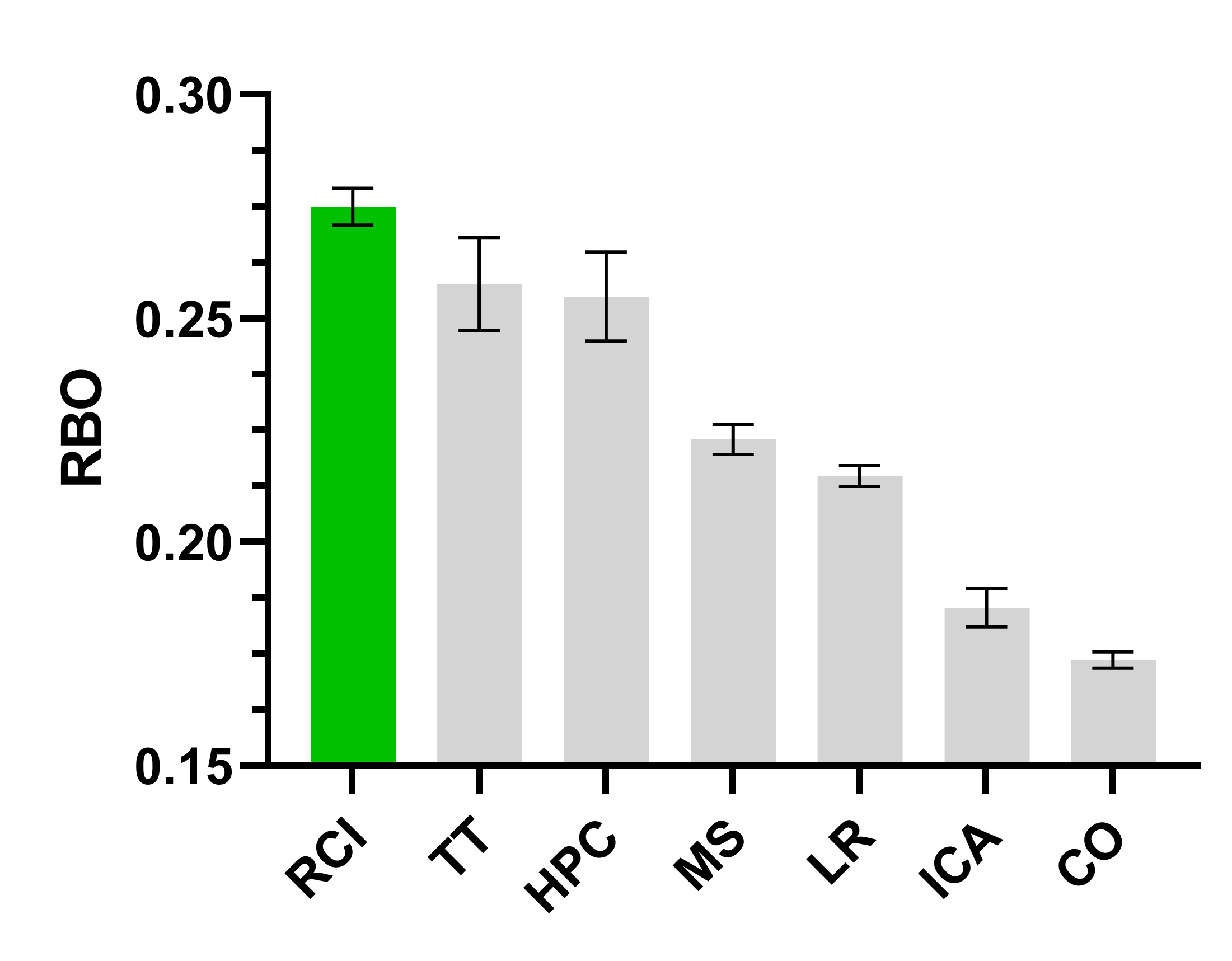}
\caption{} \label{fig_real:PBC_RBO}
\end{subfigure}
\begin{subfigure}{0.23\textwidth}
\centering
\includegraphics[scale=0.5]{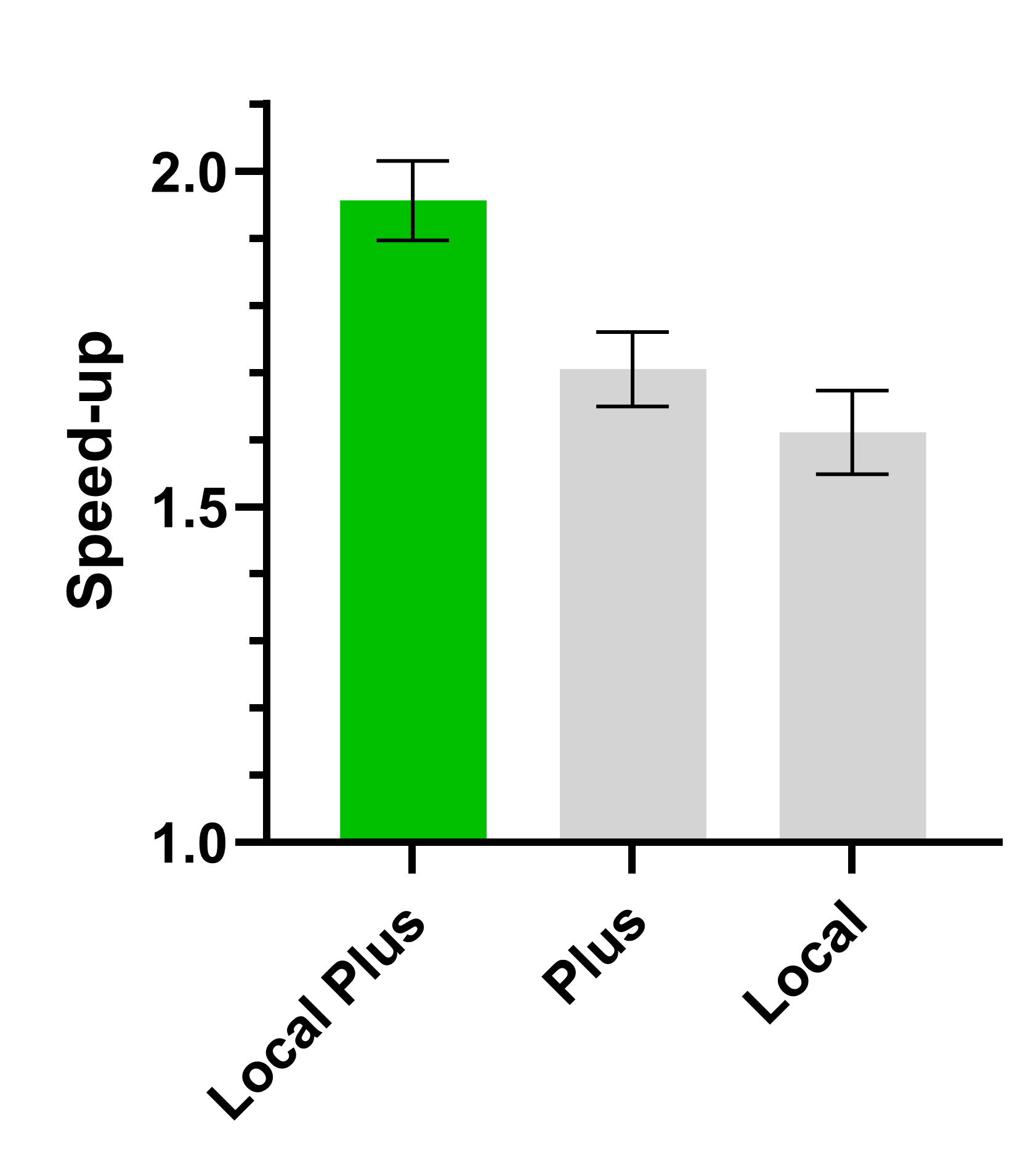}
\caption{} \label{fig_real:PBC_time}
\end{subfigure}
\hspace{10mm}\begin{subfigure}{0.23\textwidth}
\centering
\includegraphics[scale=0.5]{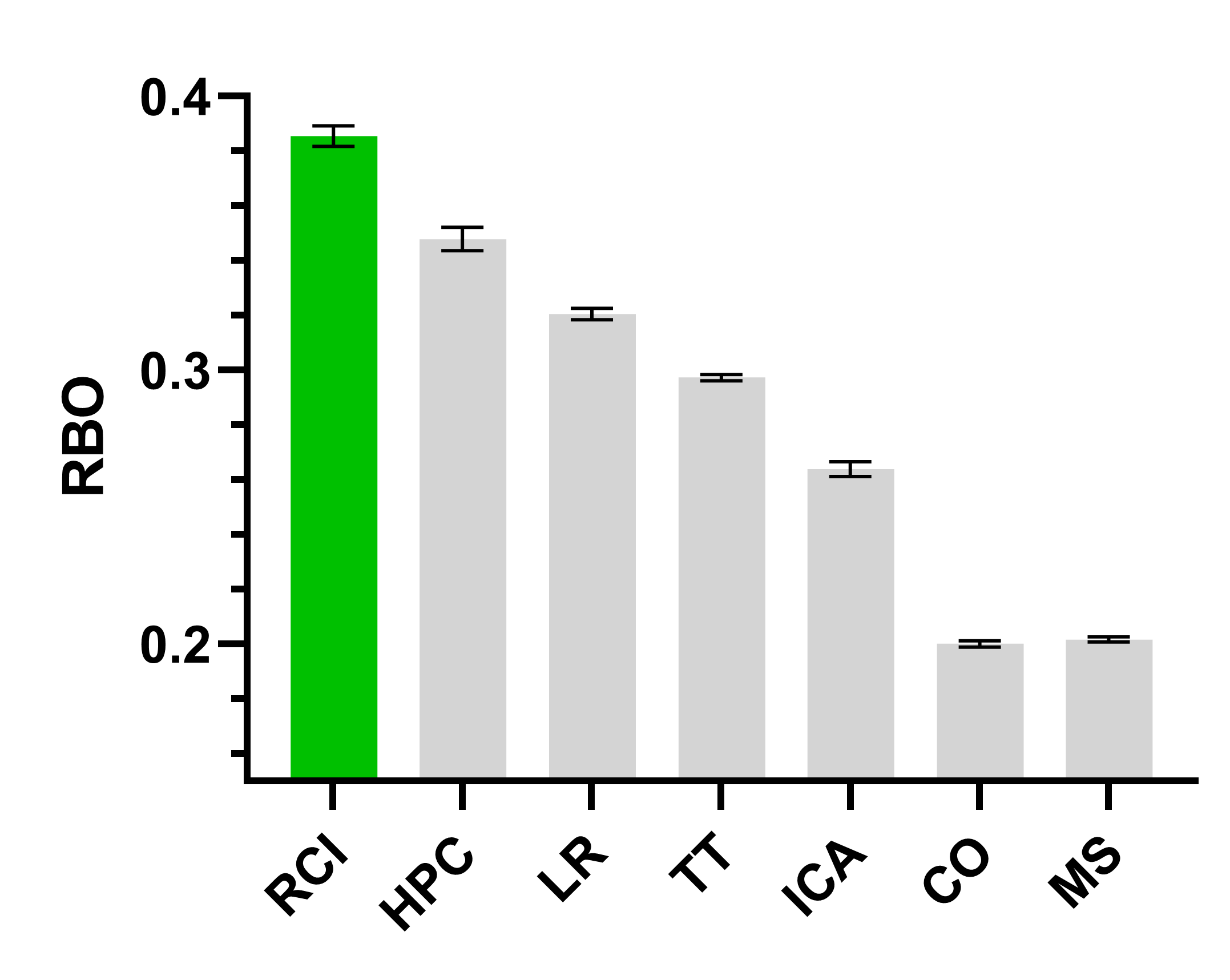}
\caption{} \label{fig_real:FHS_RBO}
\end{subfigure}
\begin{subfigure}{0.23\textwidth}
\centering
\includegraphics[scale=0.5]{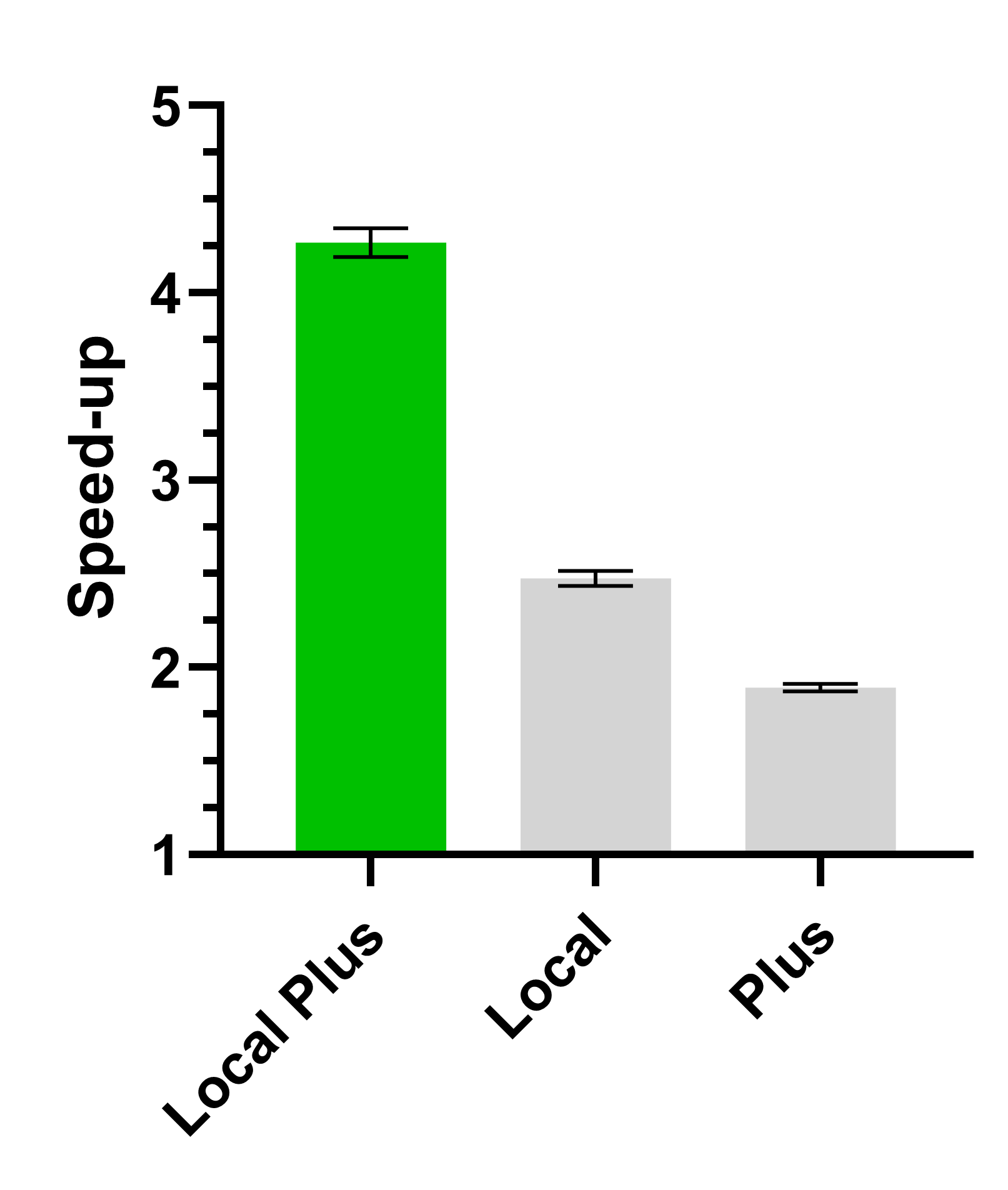}
\caption{} \label{fig_real:FHS_Time}
\end{subfigure}
\caption{Results with real data sorted from best to worst. Error bars denote 95\% confidence intervals. Sub-figures (a) and (b) correspond to accuracy and timing, respectively, for PBC. Likewise for (c) and (d) for FHS. Higher is better in each case.}
\end{figure*}

We want to identify the patient-specific root causes of mortality. Of the variables given, age and bilirubin cause death because older patients pass away and increased bilirubin levels cause neurotoxicity \cite{Lopez14}. The other variables are likely non-ancestors of death, since intervening on these variables has not been consistently shown to decrease mortality. High bilirubin levels increase the risk of death much more than age. We therefore set the gold standard ranking as age then bilirubin, if bilirubin is less than 2 mg/dL in accordance with the classic Child-Turcotte cut-off, and bilirubin then age otherwise \cite{Child64}.   

We ran all of the algorithms on 1000 bootstrapped draws of the dataset. We report accuracy results in Figure \ref{fig_real:PBC_RBO}. RCI achieved the best accuracy, followed by HPC and TT. The results therefore mimic those seen with the synthetic data. Ablation studies in Figure \ref{fig_real:PBC_time}  also show that Local Plus sped up the original version of DL by around a factor of two. We conclude that RCI quickly identifies the root causes of mortality with the highest degree of accuracy in this dataset.

\subsubsection{Framingham Heart Study}

We can also assess performance by using time information, even if we do not know the root causes for each patient. The Framingham Heart Study (FHS) contains a longitudinal dataset of 2031 patients with three time steps of clinical variables related to the cardiovascular system: systolic blood pressure, diastolic blood pressure, heart rate, total cholesterol, BMI, glucose \cite{Dawber51}. The first time step also includes age, and the third HDL and LDL. 

We want to identify the patient-specific root causes of overweightness or obesity, defined as BMI $\geq$ 25, in the first time step. Causes must precede the diagnosis in time, so the top variables for the ground truth ranking can only include the variables in the first time step. We rank the first time step variables by their z-score for each patient, since we do not know which variables have a stronger causal relation to BMI on a per-person basis. 

We run all of the algorithms without time step information. We then assess how often the algorithms rank variables in the first time step in the top spots by again using the RBO metric. We report results over 1000 bootstrapped datasets in Figure \ref{fig_real:FHS_RBO}. RCI again achieved the highest mean RBO value, this time outperforming the other algorithms by an even wider margin. Moreover, Local Plus sped-up DL by a factor of four per Figure \ref{fig_real:FHS_Time}. Both of these results are consistent with the synthetic data, where RCI continues to improve with higher sample sizes, and Local Plus completes sooner with more dimensions. We conclude that RCI and Local Plus again perform the best with the FHS dataset.

\section{Conclusion}
We described patient-specific root causes as variables subject to exogenous ``shocks'' that then predictably induce disease. In other words, the variables are associated with the predictive exogenous error terms of a structural equation model. We quantified predictivity using sample-specific Shapley values of the log-odds. This formulation allowed us to develop the RCI algorithm which recovers patient-specific Shapley values using the independent errors extracted from an SEM obeying the LiNGAM model. We then customized a critical sub-procedure of RCI in order to speed up the algorithm. Experiments highlighted the superior accuracy and speed of RCI in identifying patient-specific root causes using both synthetic and real data. See the Supplementary Materials for a discussion about limitations.

\bibliographystyle{ACM-Reference-Format}
\bibliography{biblio}

\section*{Supplementary Materials}

\subsection*{Additional Background}

The triple $\langle Z_i, Z_j, Z_k \rangle$ forms a \textit{collider} in a directed graph $\mathbb{G}$ if $Z_i \rightarrow Z_j \leftarrow Z_k$, and $Z_i$ and $Z_k$ are non-adjacent. $Z_i$ and $Z_j$ are \textit{d-connected} given $\bm{W} \subseteq \bm{Z} \setminus \{Z_i, Z_j\}$ in $\mathbb{G}$, if there exists a path between $Z_i$ and $Z_j$ such that any collider on the path is an ancestor of $\bm{W}$ and no non-collider on the path is in $\bm{W}$. On the other hand, $Z_i$ and $Z_j$ are \textit{d-separated} given $\bm{W}$ if they are not d-connected given $\bm{W}$.

We can associate a density $p(\bm{Z})$ to a DAG $\mathbb{G}$ by requiring that it factorize into the product of conditional densities of each variable given its parents:
\begin{equation} \nonumber
p(\bm{Z})=\prod_{i=1}^{p+1} p(Z_i | \textnormal{Pa}_{\mathbb{G}}(Z_i)).
\end{equation}
Any distribution which factorizes according to the above equation also satisfies the \textit{global Markov property} where, if $Z_i$ and $Z_j$ are d-separated given $\bm{W}$ in $\mathbb{G}$, then $Z_i$ and $Z_j$ are conditionally independent given $\bm{W}$ \cite{Lauritzen90}. We refer to the converse of the global Markov property as \textit{d-separation faithfulness}.

\subsection*{Proofs}
\begin{replemma}{lem_anc}
If $D$ is a terminal vertex, then $D \ci E_i | \bm{W}$ for any $E_i \not \in \textnormal{Anc}_{\mathbb{G}^\prime}(D)$ and $\bm{W} \subseteq (\bm{E} \setminus E_i)$.
\end{replemma}
\begin{proof}
Consider the augmented graph $\mathbb{G}^\prime$. The density over $\bm{Y} = \bm{Z} \cup \bm{E}$ factorizes as follows:
\begin{equation} \nonumber
    p(\bm{Y}) = \prod_{i=1}^{2(p+1)} p(Y_i | \textnormal{Pa}_{\mathbb{G}^\prime}(Y_i)).
\end{equation}
Note that $D$ and $E_i$ are d-separated given $\bm{W}$ in $\mathbb{G}^\prime$ for any $E_i \not \in \textnormal{Anc}_{\mathbb{G}^\prime}(D)$. The conclusion follows by the global Markov property.
\end{proof}

\begin{reptheorem}{thm_RCI}
Under Assumptions \ref{assump_LiNGAM} and \ref{assump_terminal}, RCI outputs the true Shapley values $\mathcal{S}_{k\cdot}$ for any patient $k$ in $\mathcal{T}$ in the large sample limit. Moreover, $s_i^k > 0$ \underline{if and only if} $X_i$ is a root cause of $D$ for patient $k$.
\end{reptheorem}
\begin{proof}
DL returns $\bm{E}_{1:p}$ in the large sample limit under Assumption \ref{assump_LiNGAM} (Lemma 1 in \cite{Shimizu11}). Logistic regression also recovers $\delta$ under Assumption \ref{assump_terminal} in the sample limit. Proposition \ref{prop_delta} ensures that 
$s_i^k = e_i^k \delta_i$ for each patient $k$ in $\mathcal{T}$. 

For the second statement, Lemma \ref{lem_anc} ensures that $s_i^k=0$ for any $E_i \not \in \textnormal{Anc}_{\mathbb{G}^\prime}(D)$. Hence, $s_i^k > 0$ implies $E_i \in \textnormal{Anc}_{\mathbb{G}^\prime}(D)$. The conclusion follows by Definition \ref{def_root}. The backward direction follows directly by Definition \ref{def_root}.
\end{proof}

\begin{repproposition}{prop_root}
 FindRoot+ and FindRoot always output the same root variable $G$.
 \end{repproposition}
 \begin{proof}
 The proof follows directly from the fact that $\textnormal{min}(0,C_{ij})^2\\ \geq 0$, so $T_i$ at any iteration is bounded below by its current score for any $i,j \in [|\bm{U}|]$.
 \end{proof}
 
 \begin{replemma}{lem_error}
Consider any $E_i \in \bm{E} \setminus E_{p+1}$. Under d-separation faithfulness and Assumption \ref{assump_terminal}, $E_i \in \textnormal{Anc}_{\mathbb{G}^\prime}(D)$ \underline{if and only if} $E_i \not \ci D$.
\end{replemma}
\begin{proof}
The backward direction follows by Lemma \ref{lem_anc}. For the forward direction, if there exists a directed path from $E_i \in \textnormal{Anc}_{\mathbb{G}^\prime}(D)$ to $D$, then $E_i \not \ci_d D$. This implies $E_i \not \ci D$ by d-separation faithfulness.
\end{proof}

\begin{reptheorem}{thm_LP}
Under d-separation faithfulness and Assumptions \ref{assump_LiNGAM} and \ref{assump_terminal}, Local Plus \underline{exclusively} recovers the error terms in $\textnormal{Anc}_{\mathbb{G}^\prime}(D) \setminus E_{p+1}$ for any $0 < \alpha \leq 1$.
\end{reptheorem}
\begin{proof}
We prove the statement by induction. Base: suppose the statement is true over any LiNGAM model where $\bm{X}$ contains only one variable. Then $X_1$ is exogenous and is either (1) left in $\bm{U}$ when $X_1 \in \textnormal{Anc}_{\mathbb{G}}(D)$, or (2) removed from $\bm{U}$ when $X_1 \not \in \textnormal{Anc}_{\mathbb{G}}(D)$ by Lines 4-9 and Lemma \ref{lem_error}. The conclusion then follows after one iteration of the repeat loop. 

Induction: Assume that the statement is true over any LiNGAM model where $\bm{X}$ contains $p$ variables. We need to prove the statement over any LiNGAM model where $\bm{X}$ contains $p+1$ variables. Without loss of generality, assume that the variables in $\bm{X}$ are permuted so that $\theta$ is strictly upper triangular with exogenous variable $X_1$. Thus $X_1$ is again either (1) left in $\bm{U}$ when $X_1 \in \textnormal{Anc}_{\mathbb{G}}(D)$, or (2) removed from $\bm{U}$ when $X_1 \not \in \textnormal{Anc}_{\mathbb{G}}(D)$ by Lines 4-9 and Lemma \ref{lem_error}. Note that $A=(I-\theta)^{-1}$ is also upper triangular with diagonal elements equal to one. Partialing out $X_1$, when it is left in $\bm{U}$, converts the first row of $A$ into a zero vector. Deleting the first row and the first column of $A$ still results in a upper triangular matrix when either (1) or (2) holds -- thus generating another LiNGAM model over $p$ variables. The conclusion follows by the inductive hypothesis. 
\end{proof}

\subsection*{Additional Experimental Results} \label{app_exp}

We report timing results for Local Plus up to $p=1000$ in Table \ref{table_sec}. Absolute times for the other variants can be derived in conjunction with Table \ref{exp_synth:time}. Local Plus extracts the error terms with $n=10,000$ and $p=1000$ in around 25 minutes.
\begin{table}[h]
\begin{tabular}{lcccc}
\hhline{=====}
        & \textit{p}=10   & \textit{p}=50   & \textit{p}=100  & \textit{p}=1,000   \\\hhline{=====}
\textit{n}=100   & 0.0027 & 0.0163 & 0.0558 & 4.627    \\
\textit{n}=1,000  & 0.0084 & 0.1349 & 0.4241 & 94.586   \\
\textit{n}=10,000 & 0.0984 & 1.2793 & 4.9641 & 1522.391 \\ \hline
\end{tabular}
\caption{Average time in seconds for Local Plus.} \label{table_sec}
\end{table}

\subsection*{Limitations}

RCI unfortunately carries several limitations. First, it only considers changes to the value of $E_i$ but not to the function $g_i$ in Equation \eqref{eq_err_change}. The framework therefore implicitly assumes that we can restore a patient to normalcy by only changing the values of $\bm{X}$. RCI may fail to detect root causes induced by, for example, \textit{unmeasured} structural changes in proteins that modify molecular interactions. The algorithm also requires unconfoundedness, acyclicity and linearity which likely do not hold with real data. Finally, RCI assumes that $D$ is a terminal vertex, but a diagnosis can disrupt molecular pathways by, for instance, causing clinicians to perform medical interventions even prior to data collection. Algorithms which drop some of the aforementioned assumptions will likely improve performance substantially.

\end{document}